\documentclass{article}

\PassOptionsToPackage{numbers, compress}{natbib}
\usepackage[final]{neurips_2024}

\usepackage[utf8]{inputenc} % allow utf-8 input
\usepackage[T1]{fontenc} % use 8-bit T1 fonts
\usepackage{hyperref} % hyperlinks
\usepackage{url} % simple URL typesetting
\usepackage{booktabs} % professional-quality tables
\usepackage{amsfonts} % blackboard math symbols
\usepackage{nicefrac} % compact symbols for 1/2, etc.
\usepackage{microtype} % microtypography
\usepackage[dvipsnames]{xcolor} % colors
\usepackage[misc]{ifsym} % for the envelope symbol
\usepackage{bm}
\usepackage{amsmath, amsfonts}
\usepackage{multirow}
\usepackage{colortbl}
\definecolor{Color-p}{RGB}{216, 239, 248}
\usepackage{wrapfig}
\usepackage{pifont}

\newcommand{\name}[1]{\textcolor{Black}{#1}}

\usepackage{amsmath,amsfonts,bm}
\usepackage{xparse}
\usepackage{cleveref}
\usepackage{environ}
\NewEnviron{smallalign}{%
\begin{align}
\scalebox{0.99}{$\BODY$}
\end{align}
}

\hypersetup{
    colorlinks=true,
    linkcolor=red,
    citecolor=cyan,
    filecolor=magenta,      
    urlcolor=cyan,
    }

\Crefformat{section}{\textcolor{red}{#2Section~#1#3}}

 \makeatother
\usepackage{algorithm}
\usepackage[noend]{algpseudocode}

\def\eqref#1{equation~\ref{#1}}
\def\1{\bm{1}}

\def\vs{{\bm{s}}}

\def\vx{{\bm{x}}}
\def\vX{{\bm{X}}}
\def\vy{{\bm{y}}}

\def\vxi{{\bm{\xi}}}

\DeclareMathAlphabet{\mathsfit}{\encodingdefault}{\sfdefault}{m}{sl}
\SetMathAlphabet{\mathsfit}{bold}{\encodingdefault}{\sfdefault}{bx}{n}

\def\gD{{\mathcal{D}}}

\def\gL{{\mathcal{L}}}

\def\gN{{\mathcal{N}}}

\def\gS{{\mathcal{S}}}
\def\gT{{\mathcal{T}}}

\def\sB{{\mathbb{B}}}

\def\sR{{\mathbb{R}}}
\def\sS{{\mathbb{S}}}

\DeclareMathOperator*{\argmax}{arg\,max}
\DeclareMathOperator*{\argmin}{arg\,min}

\title{Diversity-Driven Synthesis: Enhancing Dataset Distillation through Directed Weight Adjustment}

\author{
Jiawei Du\textsuperscript{1,2}\quad
Xin Zhang\textsuperscript{1,2,3} \quad
Juncheng Hu\textsuperscript{4} \quad
Wenxing Huang\textsuperscript{1,2,5} \quad
Joey Tianyi Zhou\textsuperscript{1,2\,\Letter}\\
{\small \textsuperscript{1} Centre for Frontier AI Research (CFAR), Agency for Science, Technology and Research (A*STAR), Singapore}\\
 {\small \textsuperscript{2} Institute of High Performance Computing, Agency for Science, Technology and Research (A*STAR), Singapore}\\
\textsuperscript{3}{\small XiDian University, Xi'an, China}  \textsuperscript{4}{\small National University of Singapore, Singapore}\\
\textsuperscript{5}{\small Hubei University, WuHan, China}  \\
}

\begin{document}

\maketitle

\renewcommand{\thefootnote}{} 
\footnotetext{Email: dujiawei@u.nus.edu, joey.tianyi.zhou@gmail.com. \textsuperscript{\Letter} represents the corresponding author.}

\vspace{-1.5em}
\begin{abstract}\label{sec:abstract}
The sharp increase in data-related expenses has motivated research into condensing datasets while retaining the most informative features. Dataset distillation has thus recently come to the fore. This paradigm generates synthetic datasets that are representative enough to replace the original dataset in training a neural network. To avoid redundancy in these synthetic datasets, it is crucial that each element contains unique features and remains diverse from others during the synthesis stage. In this paper, we provide a thorough theoretical and empirical analysis of diversity within synthesized datasets. We argue that enhancing diversity can improve the parallelizable yet isolated synthesizing approach. Specifically, we introduce a novel method that employs dynamic and directed weight adjustment techniques to modulate the synthesis process, thereby maximizing the representativeness and diversity of each synthetic instance. Our method ensures that each batch of synthetic data mirrors the characteristics of a large, varying subset of the original dataset. Extensive experiments across multiple datasets, including CIFAR, Tiny-ImageNet, and ImageNet-1K, demonstrate the superior performance of our method, highlighting its effectiveness in producing diverse and representative synthetic datasets with minimal computational expense. Our code is available at \href{https://github.com/AngusDujw/Diversity-Driven-Synthesis}{https://github.com/AngusDujw/Diversity-Driven-Synthesis}.

\end{abstract}
\vspace{-1.5em}
\section{Introduction}\label{sec:Introduction}
% Condensing datasets while preserving their key characteristics is a significant challenge in machine learning. With the rapid growth in dataset size and the need for efficient data storage and processing~\cite{kuznetsova2020open, gao2020pile}, dataset distillation has received increasing attention in recent years.

With the rapid growth in dataset size and the need for efficient data storage and processing~\cite{gao2020pile, kuznetsova2020open, kaplan2020scaling, hoffmann2022training}, how to condense datasets while preserving their key characteristics becomes a significant challenge in machine learning community~\cite{he2024efficient, tirumala2023d}. Unlike previous research~\cite{paul2021deep, toneva2018an, zhang2024TDDS, xu2024efficient} that focuses on constructing a representative subset through selecting from the original data, \emph{Dataset Distillation}~\cite{DD, sachdeva2023data, lei2023comprehensive} aims to synthesize a small and compact dataset that retains informative features from the original dataset. A model trained on the synthetic dataset is thus supposed to achieve comparable performance as one trained on the original dataset. The development of dataset distillation reduces data-related costs~\cite{feng2023embarrassingly, shang2023mimdd, zhang2023accelerating} and helps us better understand how Deep Neural Networks (DNNs) extract knowledge from large-scale datasets.

\begin{wrapfigure}{L}{0.6\textwidth}
 \centering
 \vspace{-1em}
 \includegraphics[width = 0.55\textwidth]{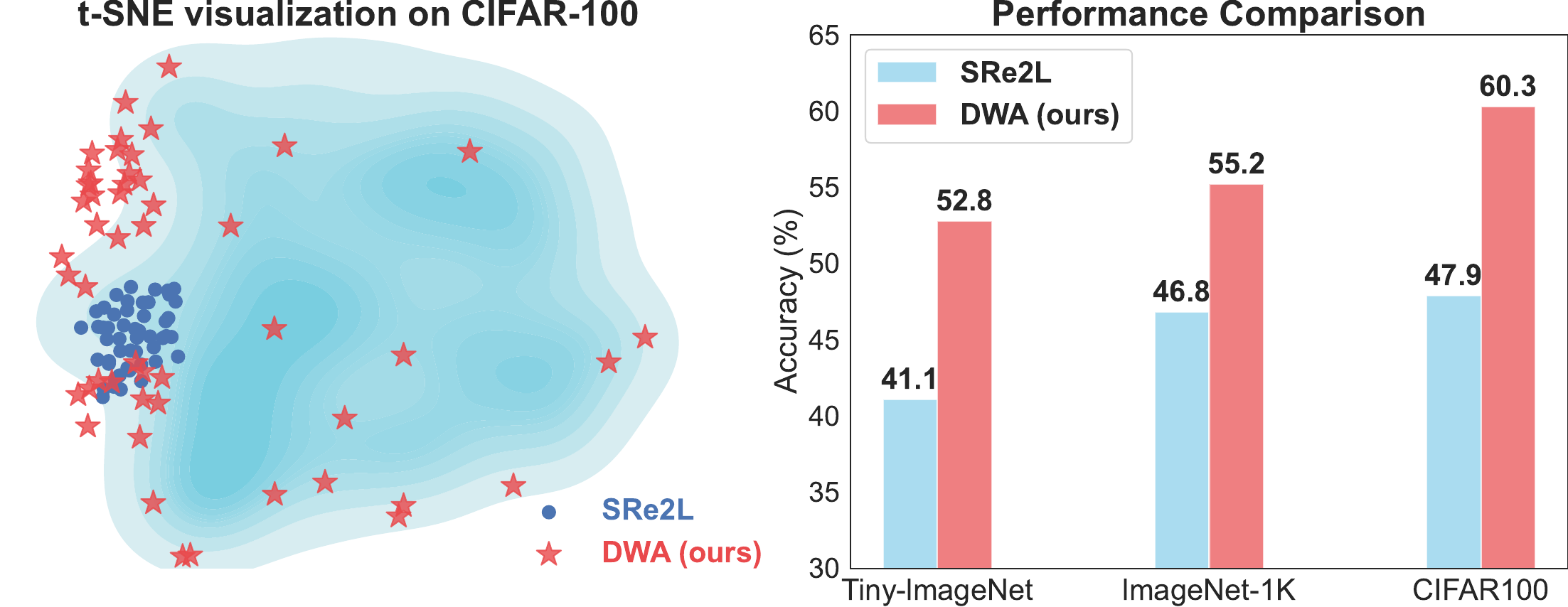}
 \caption{\small \textbf{Left:} t-SNE visualization of logit embeddings on CIFAR-100~\cite{cifar} dataset. The scatter plot illustrates the distribution of synthetic data instances distilled by SRe2L (blue dots) and our DWA method (red stars). The blue density contours represent the distribution of natural data instances. Our DWA method demonstrates a more diverse and widespread distribution compared to SRe2L~\cite{sre}, indicating better generalization and coverage of the feature space. \textbf{Right:} The consequent performance improvement of \name{DWA} in various datasets. Experiments are conducted with 50 images per class.}
 \label{fig:intro}
 \vspace{-1em}
\end{wrapfigure}

Numerous studies dedicate significant effort to synthesizing distilled datasets more effectively. For example, Zhao \textit{et al.} employ a gradient-matching approach~\cite{zhao2021dataset, DC} to guide the synthesis process. Trajectory-matching methods~\cite{MTT, tesla, FTD, du2024sequential} further align gradient trajectories to optimize the synthetic data. Additionally, distribution matching~\cite{CAFE, DM, zhao2023improved} and kernel inducing points methods~\cite{KIP, rfad, liu2023dream,loo2023dataset} also contribute to synthesizing representative data. Despite the great progress achieved by these methods on datasets like CIFAR~\cite{cifar}, their extensive computational overhead (both GPU memory and GPU time) hinders the extension of these methods to large-scale datasets like ImageNet-1K~\cite{Imagenet}.

Several recent works~\cite{tesla, sre, liu2024mgdd,break2024,zhou2024self} have attempted to address the efficiency issues of dataset distillation. In particular, Yin \textit{et al.}~\cite{sre} propose a lightweight distillation method, \name{SRe2L}, which successfully condenses the large-scale dataset ImageNet-1K. Unlike previous methods~\cite{MTT, DM, kim2022dataset} that treat the synthetic set as a unified entity to utilize the mutual influences among synthetic instances, \name{SRe2L} synthesizes each synthetic data instance individually. As such, \name{SRe2L} significantly reduces both GPU memory costs and computational overhead.

Individually synthesizing each data instance can efficiently parallelize optimization tasks, thereby flexibly managing GPU memory usage and computational overhead. However, this approach may present challenges in ensuring the representativeness and diversity of each instance. If each instance is synthesized in isolation, there may be a risk of missing the holistic view of the data characteristics, which is crucial for the training of generalized neural networks. Intuitively, \name{SRe2L} might expect that random initialization of synthetic data would provide sufficient diversity to prevent homogeneity in the synthetic dataset. Nevertheless, our analysis, as demonstrated in \autoref{fig:intro}, reveals that this initialization contributes only marginally to diversity. Conversely, the Batch Normalization (BN) loss~\cite{yin2020dreaming} in SRe2L plays the practical role in enhancing diversity of the distilled dataset.

Motivated by these findings, we further investigate the factors that enhance the diversity of synthetic datasets from a theoretical perspective. We reveal that the variance regularizer in the BN loss is the key factor ensuring diversity. Conversely, the mean regularizer within the same BN loss unexpectedly constrains diversity. To resolve this contradiction, we suggest a decoupled coefficient to specifically strengthen the variance regularizer’s role in promoting diversity. Experimental results validate our hypothesis. We further propose a dynamic mechanism to adjust the weight parameters of the teacher model. Serving as the sole source of supervision from the original dataset, the teacher model guides the synthesis comprehensively. Our meticulously designed weight perturbation mechanism injects randomness without compromising the informative supervision, thereby improving overall performance. Importantly, our method incurs negligible additional computations ($< 0.1\%$). Intuitively, our method perturbs the weight in a direction that reflects the characteristics of a large subset, varying with each batch of synthesized data.

We conduct extensive experiments across various datasets, including CIFAR-10, CIFAR-100, Tiny-ImageNet, and ImageNet-1K, to verify the effectiveness of our proposed method. The superior performance of our method not only validates our hypothesis but also demonstrates its ability to enhance the diversity of synthetic datasets. This success guides further investigations into searching for representative synthetic datasets for lossless dataset distillation.
Our contribution can be summarized as follows:
\begin{itemize}
\item We analyze the diversity of the synthetic dataset in dataset distillation both theoretically and empirically, identifying the importance of ensuring diversity in isolated synthesizing approaches.
\item We propose a dynamic adjustment mechanism to enhance the diversity of the synthesized dataset, incurring negligible additional computations while significantly improving overall performance. Extensive experiments on various datasets verify the remarkable performance of our method.
\end{itemize} 
% \vspace{-1em}
\section{Preliminaries}
\textbf{Notation and Objective.}
Given a real and large dataset $\gT = \{ (\tilde{\vx}_i, \vy_i)\}_{i = 1}^{|\gT|}$, Dataset Distillation aims to synthesize a tiny and compact dataset $\gS = \{ (\tilde{\vs}_i, \vy_i)\}_{i = 1}^{|\gS|}$. The samples in $\gT$ are drawn i.i.d from a natural distribution $\gD$, while the samples in $\gS$ are optimized from scratch. We use $\theta_{\gT}$ and $\theta_{\gS}$ to represent the converged weight trained on $\gT$ and $\gS$, respectively. We define a neural network $h = g \circ f$, where $g$ acts as the feature extractor and $f$ as the classifier. The feature extractor and the classifier loaded with the corresponding weight parameters from $\theta$ are denoted by $g_\theta$ and $f_\theta$. 

Throughout the paper, we explore the properties of synthesized datasets within the latent space. We transform both $\tilde{\vx}, \tilde{\vs} \in \sR^{\mathrm{C} \times \mathrm{H} \times \mathrm{W}}$ from the pixel space, to the latent space, $\vx, \vs \in \sR^d$, for better formulation. This transformation is given by $\vx = g_{\theta_{\gT}} (\tilde{\vx})$ and $\vs = g_{\theta_{\gT}} (\tilde{\vs})$. The objective of Dataset Distillation is to ensure that a model $h$ trained on the synthetic dataset $\gS$ is able to achieve a comparable test performance as the model trained with $\gT$, which can be formulated as, 
\begin{equation}
\label{eq:dd_objective}
 	\mathop{\mathbb{E}}_{\vx \sim \gD} \left[\ell \left(h_{\theta_{\gT}}, \vx \right) \right] \simeq \mathop{\mathbb{E}}_{\vx \sim \gD} \left[\ell \left(h_{\theta_{\gS}}, \vx \right) \right], 
\end{equation}
where $\ell$ can be an arbitrary loss function. The expression $\ell (h_{\theta_{\gT}}, \vx)$ should be interpreted as $\ell (h_{\theta_{\gT}}, \vx, \vy)$, where $\vy$ is the ground truth label.
 
\textbf{Synthesizing $\gS$.}
A series of previous works mentioned in \Cref{sec:related} have introduced various methods to synthesize $\gS$. Specifically, \name{SRe2L}~\cite{sre} proposes an efficient and effective synthesizing method, which optimizes each synthetic instance $\vs_i$ by solving the following minimization problem\footnote{In the actual optimization process, operations occur within the pixel space using the entire network $h_{\theta_{\gT}}$. However, as we discuss the optimization in the latent space, we only consider solutions within this space. Then, we transform the solution in latent space back into pixel space as $\tilde{\vs} = g_{\theta_{\gT}}^{-1} (\vs)$.}:
\begin{equation}
\label{eq:srel_obj}
	\argmin_{\vs_i \in \sR^d} \left[\ell \left(f_{\theta_{\gT}}, \vs_i \right) + \lambda \gL_\mathrm{BN} \left(f_{\theta_{\gT}}, \vs_i \right) \right], 
\end{equation}
where $\gL_\mathrm{BN}$ denotes the BN loss, and $\lambda$ is the coefficient of $\gL_\mathrm{BN}$. The detailed definition of $\gL_\mathrm{BN}$ can be found in \autoref{eq:BN_defi}. Minimizing the BN loss $\gL_\mathrm{BN}$ significantly enhances the performance of \name{SRe2L}, which is designed to ensure that $\gS$ aligns with the same normalization distribution as $\gT$. However, we argue that another essential but overlooked aspect of the BN loss $\gL_\mathrm{BN}$ is its role in introducing diversity to $\gS$, which also greatly benefits the final performance. In the following section, we will analyze this issue in greater detail.

%Minimizing $\ell (f_{\theta_{\gT}}, \vs_i)$ ensures the synthesize data instance $s_i$ to contains intra-class features; while minimizing $\gL_\mathrm{BN}$ is claimed by 

%Intuitively, the synthesize data instance $s_i$ is composed of condensed intra-class features. Meanwhile, $s_i$ is designed to adhere to the established normalization standards of $f_{\theta_{\gT}}$, ensuring that its values fall within the same normalization distribution as those in $\gT$.

%Unlike other approaches that consider the mutual influences among synthetic data instance and optimize the dataset collectively, \name{SRe2L} optimizes each synthetic data instance individually. As shown in \autoref{eq:srel_obj}, solving this minimization problem potentially yields many homogeneous solutions. The only factor to ensure the diversity of the synthetic dataset $\gS$ is the initialization of each synthetic instance. In the following section, we will analyze this issue in greater detail. 

\section{Methodology}\label{sec:Methodology}
%Ensuring the diversity of the synthetic dataset $\gS$ is critical for utilizing the limited distillation budget effectively. In this section, we first reveal that the effectiveness of the BN loss, as referred to in \autoref{eq:srel_obj}, is also due to its contribution to enhancing the diversity of $\gS$. Then, we argue that while the BN loss is the only factor in ensuring diversity, it is not set optimally, leading to a lack of diversity. To address this limitation, we propose integrating a dynamic adjustment mechanism into the weight parameters of $f_{\theta_{\gT}}$ to maintain and enhance diversity throughout the synthesizing phase. Finally, we present our complete algorithm and provide a theoretical demonstration of its effectiveness.
Diversity in the synthetic dataset $\gS$ is essential for effective use of the limited distillation budget. This section reveals that the BN loss, referenced in \autoref{eq:srel_obj}, enhances $\gS$'s diversity. However, the suboptimal setting of BN loss limits this diversity. To overcome this, we propose a dynamic adjustment mechanism for the weight parameters of $f_{\theta_{\gT}}$, enhancing diversity during synthesis. Finally, we detail our algorithm and theoretically demonstrate its effectiveness. The pseudocode of our proposed DWA can be found in \Cref{alg:dwa}. 

\begin{algorithm}[H] 
\caption{\small Directed Weight Adjustment (DWA)}
\thispagestyle{empty}
\label{alg:dwa}
\begin{algorithmic}[1] 
\Require Original dataset $\gT$; 
Number of iterations $T$; Image per class $\texttt{ipc}$; Number of steps $K$, magnitude $\rho$ to solve the weight adjustment $\widetilde{\Delta\theta}$; Learning rate $\eta$; A network $f_{\theta_\gT}$ with weight parameter $\theta_\gT$, $f_{\theta_\gT}$ is well trained on $\gT$.
\State Initialize $\gS = \{\}$, $\Delta\theta_0 = \mathbf{0}_{\mathrm{dim} (\theta_{\gT})}$
%\For{$t = 1$ to $T$} 
\For{$i = 1$ to $\texttt{ipc}$} 
 % \State Initialize subset $\gP^k = \{\}$, the UFCs set $\gU^k = \{\}$, and the static labels set $\gY^k = \{\}$
\State Randomly select one instance for each class from $\gT$, to initialize $\gS^i_0$, \textit{i.e.}, 
\State $\gS^i_0 = \{ (\vx_i, \vy_i) \mid (\vx_i, \vy_i) \in \gT \text{ and each } \vy_i \text{ is unique}\}$
%\State Initialize $\gU^k$ with zeros, where each $\vu_i$ has the same dimensions as $\vx$:
\State $\triangleright$ Compute the adjustment of weights $\Delta\theta$ by solving \autoref{eq:delta_theta}
\For{$k = 1$ to $K$}
\State $\Delta\theta_k$ = $\Delta\theta_{k-1}$ + $\frac{\rho}{K} \nabla L_{\gS^i_0} (f_{\theta_\gT + \Delta\theta_{k-1}})$ \label{eq:solve_delta_theta}
\EndFor
\State $\widetilde{\Delta\theta} = \Delta\theta_K$ \Comment{Directed Weight Adjustment}
\State $\triangleright$ Optimize $\gS^i$
\For{$t = 1$ to $T$}
\State $\gS^i_{t} = \gS^i_{t-1} + \eta \nabla_{\gS} \gL (f_{\theta_\gT + \widetilde{\Delta\theta}}, \gS^i_{t-1})$ \Comment{$\gL$ is defined in \autoref{eq:final_objective}}
%\State Update $\gS^k_{t}$ with $f_{\theta_\gT + \widetilde{\Delta\theta}}$ by 
\EndFor
\State $\gS = \gS \cup \{\gS^i\}$
\EndFor
	\Ensure Synthetic dataset $\gS$
	\end{algorithmic}
\end{algorithm}

%The primary limitation in synthesizing a distilled dataset is the homogeneity of the synthetic data instances. Solving the minimization problem outlined in \autoref{eq:srel_obj} often overlooks the mutual influences among synthetic data instances, typically resulting in many homogeneous solutions. In this section, we introduce a novel and efficient methodology designed to enhance both the diversity and quality of synthetic data instances. Initially, we address the lack of diversity limitation in synthesizing the distilled dataset. Subsequently, we integrate a dynamic adjustment mechanism into the weight parameters of $f_{\theta_{\gT}}$ to maintain and enhance diversity throughout the synthesizing phase. Finally, we present our complete algorithm and provide a theoretical demonstration of its effectiveness.

%In this section, we introduce a novel and efficient methodology designed to optimize the diversity and quality of synthetic data instances. Initially, we address the lack of diversity, a limitation highlighted in preliminary section. We then offer a new perspective to understand the homogeneous solutions of \autoref{eq:srel_obj}. Subsequently, we integrate a dynamic adjustment mechanism into the weight parameters of $f_{\theta_{\gT}}$ to maintain and enhance diversity throughout the synthesizing phase. Finally, we present our complete algorithm and theoretically demonstrate its effectiveness.

\subsection{Batch Normalization Loss Enhances Diversity of $\gS$} \label{sec: decoupled_varaince}
The BN loss $\gL_\mathrm{BN}$ comprises mean ($\gL_\mathrm{mean}$) and variance ($\gL_\mathrm{var}$) components, defined as follows: 
%The definition of the BN loss $\gL_\mathrm{BN}$ is as below, 
\begin{align}
\label{eq:BN_defi}
	\gL_\mathrm{BN} = \gL_\mathrm{mean} + \gL_\mathrm{var} \quad \mbox{where} \quad & \gL_\mathrm{mean} \left(f_{\theta_{\gT}}, \vs_i \right) = \sum\nolimits_l \left\|\mu_l \left(\sS \right) - \mu_l \left(\gT \right) \right\|_2, \nonumber\\
	\quad \mbox{and} \quad & \gL_\mathrm{var} \left(f_{\theta_{\gT}}, \vs_i \right) = \sum\nolimits_l \left\|\sigma^2_l \left(\sS \right) -\sigma^2_l \left(\gT \right) \right\|_2, 
\end{align}
where $\mu_l$ and $\sigma^2_l$ refer to the channel mean and variance in the $l$-th layer, respectively. $\vs_i$ is optimized within a mini-batch $\sS$, where $\vs_i \in \sS$ and $\sS \subset \gS$. Each component of $\gL_\mathrm{BN}$ operates from its own perspective to enhance dataset distillation. First, the mean component $\gL_\mathrm{mean}$ regularizes the synthetic data $\vs$, ensuring its values align closely with those of the representative centroid of $\gT$ in latent space. Second, the variance component $\gL_\mathrm{var}$ encourages the synthetic data in $\sS$ to differ from each other, thereby maintaining the variance $\sigma^2_l (\sS)$. Thus, this BN loss-driven synthesis can be decoupled as\footnote{We disregard the class differences in the following analysis since they are identical across all classes.}
\begin{align}
\vs_i = \vX_c \left(\lambda\gL_\mathrm{mean}, \theta_{\gT} \right) + \vxi_i, 
 	%\nabla_{\theta} \ell (f_{\theta_{\gT}}, \vX_c) \leq \alpha_1 \quad & \mbox{and} \quad \gL_\mathrm{mean} (f_{\theta_{\gT}}, \rvx_c) = \sum\nolimits_l \|\mu_l (\vX_c) - \mu_l (\gT)\|_2 \leq \alpha_2, \nonumber
 % (\lambda\gL_\mathrm{var})
\end{align}
where $\vX_c$ can be regarded as an optimal solution to \autoref{eq:srel_obj} when the variance regularization term $\gL_\mathrm{var}$ is not considered, \textit{i.e.}, 
\begin{equation}
	\left\|\nabla_{\theta} \ell \left(f_{\theta_{\gT}}, \vX_c \right) \right\|_2 \leq \alpha_1 \quad \mbox{and} \quad \gL_\mathrm{mean} \left(f_{\theta_{\gT}}, \vX_c \right) = \sum\nolimits_l \left\|\mu_l \left(\vX_c \right) - \mu_l \left(\gT \right) \right\|_2 \leq \alpha_2, 
\end{equation}
where both $\alpha_1, \alpha_2 >0$ and $\alpha_1, \alpha_2 \rightarrow 0$. $\vxi_i$ represents a small perturbation and $\vxi_i\sim \gN \left(0, \sigma^2_{\vxi} (\lambda\gL_\mathrm{var}) \right)$. Therefore, the variance of the synthetic dataset $\gS$ is, 
\begin{equation}
\label{eq: varS}
 \mathrm{Var} (\gS) = \mathrm{Var} \big(\vX_c (\lambda\gL_\mathrm{mean}, \theta_{\gT}) \big) + \mathrm{Var} \left(\vxi \right) = \sigma^2_{\vxi} \left(\lambda\gL_\mathrm{var} \right).
\end{equation}
We have $\mathrm{Var} \big(\vX_c (\lambda\gL_\mathrm{mean}, \theta_{\gT}) \big) = 0$ as $\vX_c$ is deterministic. Unlike other approaches that consider the mutual influences among synthetic data instances and optimize the dataset collectively, \name{SRe2L}~\cite{sre} optimizes each synthetic data instance individually. Therefore, the diversity of the synthetic dataset $\gS$ is solely determined by $\lambda\gL_\mathrm{var}$. 

However, simply increasing $\lambda$ contributes marginally to enhancing the diversity of $\gS$. This is because a greater $\lambda$ will also emphasize the regularization term $\lambda\gL_\mathrm{mean}$, which contradicts the emphasis on $\lambda\gL_\mathrm{var}$. We provide a detailed analysis in the Appendix \ref{app:contradictory_mean_var}. As a result, we propose using a decoupled coefficient, $\lambda_\mathrm{var}$, to enhance the diversity of $\gS$.

%We illustrate the synthesis process in \autoref{fig:intro}, where the initialized 
Additionally, the synthetic data instances are optimized individually to approximate the representative data instance $\vX_c$. However, the gaussian initialization $\gN (0, 1)$ in pixel space does not distribute uniformly around $\vX_c$ in latent space, making the converged synthetic data instances to cluster in a crowed area in latent space, as dedicated in \autoref{fig:intro}. To address this, we propose initializing with real instances from $\gT$ inspired by \name{MTT}~\cite{MTT}, ensuring a uniform projection when synthesizing $\gS$.

\subsection{Random Perturbation on $\theta_\gT$ Helps Improve Diversity}
In the previous section, we highlighted the often overlooked aspect of the BN loss in introducing diversity to $\gS$, which was also verified through experiments in \Cref{sec:ablation}. Building upon this, we propose to introduce randomness into $\theta_\gT$ to further enhance $\gS$'s diversity, as it is the only remaining factor affecting $\mathrm{Var} (\gS)$, as shown in \autoref{eq: varS}.

Let $\vx^*_c = \vX_c (\lambda\gL_\mathrm{mean}, \theta_{\gT})$ to be the original optimal solution to \autoref{eq:srel_obj}. We aim to solve the adjusted optimal solution $\vx_c = \vX_c (\lambda\gL_\mathrm{mean}, \theta_{\gT} + \Delta\theta) = \vx^*_c + \Delta\vx$, where $\theta_\gT$ is randomly perturbed by $\Delta\theta$, and $\Delta\theta\sim\gN (0, \sigma^2_{\theta})$. Consequently, we have:
\begin{align}
\label{eq:deltax}
	\left\|\nabla_{\theta} \ell \left(f_{\theta_{\gT} + \Delta\theta}, \vx_c \right) \right\|_2 = \left\|\nabla_{\theta} \ell \left(f_{\theta_{\gT} + \Delta\theta}, \vx^*_c + \Delta\vx \right) \right\|_2 \leq \alpha_1.
\end{align}
To solve for $\Delta\vx$, we can apply a first-order bivariate Taylor series approximation because $\nabla_{\theta} \ell (f_{\theta_{\gT}}, \vX_c) \leq \alpha_1$, where $\alpha_1 \rightarrow 0$, and both $\Delta\theta$ and $\Delta\vx$ are small. Thus, 
\begin{align}
	 & \left\|\nabla_{\theta} \ell \left(f_{\theta_{\gT + \Delta\theta}}, \vx^*_c + \Delta\vx \right) \right\|_2 \nonumber\\
	 = & \left\|\nabla_{\theta} \ell \left(f_{\theta_{\gT}}, \vx^*_c \right) + \nabla^2_\theta\ell \left(f_{\theta_{\gT}}, \vx^*_c \right) \Delta\theta + \nabla_{\vx} \left[\nabla_{\theta}\ell \left(f_{\theta_{\gT}}, \vx^*_c \right) \right] \Delta\vx \right\|_2 \nonumber\\
	 \leq & \left\|\nabla_{\theta} \ell \left(f_{\theta_{\gT}}, \vx^*_c \right) \right\|_2 + \left\|\nabla^2_\theta\ell \left(f_{\theta_{\gT}}, \vx^*_c \right) \Delta\theta + \nabla_{\vx} \left[\nabla_{\theta}\ell \left(f_{\theta_{\gT}}, \vx^*_c \right) \right] \Delta\vx \right\|_2 \nonumber\\
	 \leq & \alpha_1 + \left\|\nabla^2_\theta\ell \left(f_{\theta_{\gT}}, \vx^*_c \right) \Delta\theta + \nabla_{\vx} \left[\nabla_{\theta}\ell \left(f_{\theta_{\gT}}, \vx^*_c \right) \right] \Delta\vx \right\|_2, 
\end{align}
To satisfy \autoref{eq:deltax}, we have:
\begin{align}
\label{eq: x_solution}
	\nabla^2_\theta\ell \left(f_{\theta_{\gT}}, \vx^*_c \right) \Delta\theta + \nabla_{\vx} \left[\nabla_{\theta}\ell \left(f_{\theta_{\gT}}, \vx^*_c \right) \right] \Delta\vx = \mathbf{0} & , \quad \mbox{then} \nonumber\\
	\Delta\vx = -\nabla_{\vx} \left[\nabla_{\theta}\ell \left(f_{\theta_{\gT}}, \vx^*_c \right) \right]^{-1}\nabla^2_\theta\ell \left(f_{\theta_{\gT}}, \vx^*_c\right) \Delta\theta & .
\end{align}
Intuitively, $\Delta\vx$ must compensate for the $\nabla_\theta$ incurred by introducing the random perturbation $\Delta\theta \sim \gN (0, \sigma^2_{\theta})$ on $\theta_\gT$. By \autoref{eq: x_solution}, $\mathrm{Var} (\Delta\vx) \propto \mathrm{Var} (\Delta\theta) = \sigma^2_{\theta}$, then:
\begin{align}
		 \mathrm{Var} \left(\gS' \right) & = \mathrm{Var} \big(\vX_c \left(\lambda\gL_\mathrm{mean}, \theta_{\gT} + \Delta\theta \right) \big) + \mathrm{Var} \left(\vxi \right) \nonumber\\
		 & = \mathrm{Var} \left(\vx^*_c + \Delta\vx \right) + \mathrm{Var} \left(\vxi \right) \nonumber\\
		 & = \beta \sigma^2_{\theta} + \sigma^2_{\vxi} \left(\lambda\gL_\mathrm{var} \right) \geq \sigma^2_{\vxi} \left(\lambda\gL_\mathrm{var} \right), 
\end{align}
where $\beta$ is determined by $-\nabla_{\vx}[\nabla_{\theta}\ell (f_{\theta_{\gT}}, \vx^c)]^{-1}\nabla^2_{\theta}\ell (f_{\theta_{\gT}}, \vx^c)$, as shown in \autoref{eq: x_solution}. Therefore, the variance of the new synthetic dataset $\gS'$ is greater than that of $\gS$ without perturbing $\theta_\gT$.

\subsection{Directed Weight Adjustment on $\theta_\gT$}
\label{sec:3.3}
Although perturbing $\theta_\gT$ could significantly increase the variance of the synthetic dataset $\gS$, undirected random perturbation $\Delta\theta$ can also introduce noise, which in turn degrades the performance. We aim to address this limitation by directing the random perturbation $\Delta\theta$ without introducing noise into $\gS$.
We propose to obtain directed $\Delta\theta$ by solving the following maximization problem:
\begin{equation}
\label{eq:delta_theta}
	\widetilde{\Delta\theta} = \argmax_{\Delta\theta} L_{\sB} \left(f_{\theta_{\gT} + \Delta\theta} \right) \quad \mbox{where} \quad L_{\sB} \left(f_{\theta_{\gT} + \Delta\theta} \right) = \sum_{\vx_i \in \sB}  \ell \left(f_{\theta_{\gT} + \Delta\theta}, \vx_i \right), 
\end{equation}
where $\sB \subset \gT $ represents a randomly selected subset of $\gT$, and $|\sB| \ll |\gT|$. As such, $\widetilde{\Delta\theta}$ will not introduce unanticipated noise when synthesizing $\gS$. The randomly selected $\sB$ ensures that the randomness of $\widetilde{\Delta\theta}$ continues to benefit the diversity of $\gS$. Next, we will demonstrate this theoretically.

Effective dataset distillation should provide concise and critical guidance from the original dataset $\gT$ when synthesizing the distilled dataset. Here, this guidance is introduced primarily through the converged weight parameters $\theta_{\gT}$, \textit{i.e.}, 
\begin{equation}
	\theta_{\gT} = \argmin_{\theta}L_{\gT} \left(f_{\theta_{\gT}} \right) \quad \mbox{where} \quad L_{\gT} \left(f_{\theta_{\gT}} \right) = \sum_{\vx_i \in \gT}  \ell \left(f_{\theta_{\gT}}, \vx_i \right),
\end{equation}
where $\theta_{\gT}$ contains informative features of $\gT$ because it achieves minimized training loss over $\gT$. We demonstrate that $\widetilde{\Delta\theta}$, obtained from \autoref{eq:delta_theta}, decreases the training loss computed over $\gT \setminus \sB$, which, in fact, highlights the features of $\gT \setminus \sB$. By applying a first-order Taylor expansion, we obtain: 
\begin{align}
\label{eq:TminusB}
	L_{\gT \setminus \sB} \left(f_{\theta_{\gT} + \widetilde{\Delta\theta}} \right) & \approx L_{\gT \setminus \sB} \left(f_{\theta_{\gT}} \right) + \nabla_\theta L_{\gT \setminus \sB} \left(f_{\theta_{\gT}} \right) \widetilde{\Delta\theta}. 
\end{align} 
Since $\theta_{\gT}$ is optimized until reaching a local minimum with respect to the loss function computed over the training set $\gT$, we have:
\begin{equation*}
	\nabla_\theta L_{\gT} \left(f_{\theta_{\gT}} \right) = \nabla_\theta L_{\sB} \left(f_{\theta_{\gT}} \right) + \nabla_\theta L_{\gT \setminus \sB} \left(f_{\theta_{\gT}} \right) = \mathbf{0} \quad \mbox{thus} \quad \nabla_\theta L_{\gT \setminus \sB} \left(f_{\theta_{\gT}} \right) = - \nabla_\theta L_{\sB} \left(f_{\theta_{\gT}} \right), 
\end{equation*}
where $\mathbf{0}$ is the tensor of zeros with the same dimension as $\theta_\gT$. Substitute it back into \autoref{eq:TminusB}, we have: 
\begin{align}
	 L_{\gT \setminus \sB} \left(f_{\theta_{\gT} + \widetilde{\Delta\theta}} \right) - L_{\gT \setminus \sB} \left(f_{\theta_{\gT}} \right) \approx & \nabla_\theta L_{\gT \setminus \sB} \left(f_{\theta_{\gT}} \right) \widetilde{\Delta\theta} \nonumber\\
	 = & - \nabla_\theta L_{\sB} \left(f_{\theta_{\gT}} \right) \widetilde{\Delta\theta} \nonumber\\
	\approx & - \left(L_{\sB} \left(f_{\theta_{\gT} + \widetilde{\Delta\theta}} \right) - L_{\sB} \left(f_{\theta_{\gT}} \right) \right) \leq 0, 
\end{align}
$L_{\sB} (f_{\theta_{\gT} + \widetilde{\Delta\theta}})$ will clearly be greater than $L_{\sB} (f_{\theta_{\gT}})$, as indicated by \autoref{eq:delta_theta}. Thus, we demonstrate that the directed $\widetilde{\Delta\theta}$ results in less noise and improved performance. In summary, after resolving $\widetilde{\Delta\theta}$ as in \autoref{eq:delta_theta}, our proposed method synthesizes data instance $\vs_i$ by solving:
\begin{equation}
\label{eq:final_objective}
	 \tilde{\vs_i} = \argmin_{\vs \in \sR^d} \gL \quad \mbox{where} \quad \gL = \left[\ell \left(f_{\theta_{\gT} + \widetilde{\Delta\theta}}, \vs_i \right) + \lambda \gL_\mathrm{mean} \left(f_{\theta_{\gT}}, \vs_i \right) + \lambda_\mathrm{var}\gL_\mathrm{var} \left(f_{\theta_{\gT}}, \vs_i \right) \right].
\end{equation}
\section{Experiments}
To evaluate the effectiveness of the proposed method, we have conducted extensive comparison experiments with SOTA methods on various datasets including CIFAR-10/100 ($32\times32$, 10/100 classes)~\cite{cifar}, Tiny-ImageNet ($64\times64$, 200 classes)~\cite{tiny}, and ImageNet-1K ($224\times224$, 1000 classes)~\cite{Imagenet} using diverse network architectures like ResNet-(18, 50, 101)~\cite{Resnet}, MobileNetV2~\cite{Mobilenetv2}, ShuffleNetV2~\cite{Shuffle}, EfficientNet-B0~\cite{efficientnet}, and VGGNet-16~\cite{vgg}. We conduct our experiments on the server with one Nvidia Tesla A100 40GB GPU.

\textbf{Solving $\widetilde{\Delta\theta}$.} Before we conduct our experiments, we propose to use a gradient descent approach to solve $\widetilde{\Delta\theta}$ in \autoref{eq:delta_theta}. There are two coefficients, $K$ and $\rho$, used in the gradient descent approach. $K$ represents the number of steps, and $\rho$ normalizes the magnitude of the directed weight adjustment. The details for solving $\widetilde{\Delta\theta}$ can be found in Line \autoref{eq:solve_delta_theta} of \Cref{alg:dwa}. 

\textbf{Experiment Setting.} Unless otherwise specified, we default to using ResNet-18 as the backbone for distillation. For ImageNet-1K, we use the pre-trained model provided by Torchvision while for CIFAR-10/100 and Tiny-ImageNet, we modify the original architecture under the suggestion in \cite{he2020momentum}.
More detailed hyper-parameter settings can be found in \Cref{app:para_setting}. 

\textbf{Baselines and Metrics.} We conduct comparison with seven Dataset Distillation methods including DC~\cite{DC}, DM~\cite{DM}, CAFE~\cite{CAFE}, MTT~\cite{MTT}, TESLA~\cite{tesla}, SRe2L~\cite{sre}, and DataDAM~\cite{DataDAM}. 
% The technical details of these comparison baselines can be found in \Cref{app:baselines}. 
For all the considered comparison methods, we assess the quality of the distilled dataset by measuring the Top-1 classification accuracy on the original validation set using models trained on them from scratch. \colorbox{Color-p!90}{Blue cells} in all tables highlight the highest performance.

\subsection{Results \& Discussions}
%To validate the robustness and generality across varing image resolutions and class numbers, we conduct extensive experiments on various datasets including CIFAR-10/100 ($32\times32$, 10/100 classes), Tiny-ImageNet ($64\times64$, 200 classes), and ImageNet-1K ($224\times224$, 1000 classes).
\textbf{CIFAR-10/100.} As shown in \autoref{tab:cifar}, our DWA exhibits superior performance compared to conventional dataset distillation methods, particularly evident on CIFAR-100 with a larger distillation budget. For instance, our DWA yields over a 10\% performance enhancement compared to MTT~\cite{MTT} with $\texttt{ipc} = 50$. Leveraging a more robust distillation backbone like ResNet-18, our approach surpasses the SOTA method SRe2L~\cite{sre} across all considered settings. Specifically, we achieve more than 5\% and 8\% accuracy improvement on CIFAR-10 and CIFAR-100, respectively.

\begin{table}[h] 
	 \centering
	 \caption{\small Comparison with SOTA dataset distillation baselines on CIFAR-10/100. Unless otherwise specified, we use the same network architecture for distillation and validation. Following the settings in their original papers, DC~\cite{DC}, DM~\cite{DM}, CAFE~\cite{CAFE}, MTT~\cite{MTT}, and TESLA~\cite{tesla} use ConvNet-128 (\textit{small model}). For SRe2L~\cite{sre}, ResNet-18 (\textit{large model}) is used for synthesis and validation.} 
	 \setlength\tabcolsep{5.5pt} 
	 % \renewcommand{\arraystretch}{1.2}
	 % {\fontsize{8.5}{9}\selectfont
	 \scriptsize
	 \begin{tabular}{cccccccc|cc}
	 \toprule 
 % \midrule
	 \multirow{2}[2]{*}{Dataset} & \multirow{2}[2]{*}{$\texttt{ipc}$} & \multicolumn{6}{c}{ConvNet} & \multicolumn{2}{c}{ResNet-18}\\
	 \cmidrule(lr){3-8} \cmidrule(lr){9-10}
 & & DC~\cite{DC} & DM~\cite{DM} & CAFE~\cite{CAFE} & MTT~\cite{MTT} & TESLA~\cite{tesla} & DWA (ours) & SRe2L~\cite{sre} & DWA (ours)\\
	 \midrule
	 \multirow{2}{*}{CIFAR-10} & $10$ & $\ApmB{44.9}{0.5}$ & $\ApmB{48.9}{0.6}$ & $\ApmB{46.3}{0.6}$ & {\cellcolor{white}$\ApmB{65.4}{0.7}$} & {\cellcolor{Color-p!90}$\ApmB{66.4}{0.8}$} & {\cellcolor{white}$\ApmB{45.0}{0.4}$} & $\ApmB{27.2}{0.4}$ & {\cellcolor{Color-p!90}$\ApmB{32.6}{0.4}$}\\
 & $50$ & $\ApmB{53.9}{0.5}$ & $\ApmB{63.0}{0.4}$ & $\ApmB{55.5}{0.6}$ & $\ApmB{71.6}{0.7}$ & {\cellcolor{Color-p!90}$\ApmB{72.6}{0.7}$} & {\cellcolor{white}$\ApmB{63.3}{0.7}$} & $\ApmB{47.5}{0.5}$ & {\cellcolor{Color-p!90}$\ApmB{53.1}{0.3}$}\\
	 \midrule
	 \multirow{2}{*}{CIFAR-100} & 10
 & $\ApmB{25.2}{0.3}$ & $\ApmB{29.7}{0.3}$ & $\ApmB{27.8}{0.3}$ & $\ApmB{40.1}{0.4}$ & $\ApmB{41.7}{0.3}$ & {\cellcolor{Color-p!90}$\ApmB{47.6}{0.4}$} & $\ApmB{31.6}{0.5}$ & {\cellcolor{Color-p!90}$\ApmB{39.6}{0.6}$}\\
 & 50 & - & $\ApmB{43.6}{0.4}$ & $\ApmB{37.9}{0.3}$ & $\ApmB{47.7}{0.2}$ & $\ApmB{47.9}{0.3}$ & {\cellcolor{Color-p!90}$\ApmB{59.0}{0.1}$} & $\ApmB{52.2}{0.3}$ & {\cellcolor{Color-p!90}$\ApmB{60.9}{0.5}$}\\
 % & & Data & Data & Data & Data & Data & & & &\\
 % \midrule
	 \bottomrule 
	 \end{tabular}
% }
	 \label{tab:cifar}
\end{table}

\begin{table}[h] 
	 \centering
	 \caption{\small Comparison with SOTA dataset distillation baselines on Tiny-ImageNet and ImageNet-1K. Unless otherwise specified, we use the same network architecture for distillation and validation. Following the settings in their original papers, MTT~\cite{MTT}, and TESLA~\cite{tesla} use ConvNet-128 (\textit{small model}). For SRe2L~\cite{sre}, ResNet-18 (\textit{large model}) is used for synthesis, and the distilled dataset is evaluated on ResNet-18, 50, and 101. $\dag$ indicates MTT is performed on a 10-class subset of the full ImageNet-1K dataset.} 
 % The accurcay of ResNet-18 and ConvNet-128 on Tiny-ImageNet is xxx and xxx, respectively. The accurcay of ResNet-18 and ConvNet-128 on ImageNet-1K is xxx and xxx, respectively
	 \setlength\tabcolsep{2.3pt} 
	 \scriptsize
	 % \renewcommand{\arraystretch}{1.4}
	 % {\fontsize{7}{7}\selectfont
	 \begin{tabular}{ccccc|cc|cc|cc}
	 \toprule 
 % \midrule
	 \multirow{2}[2]{*}{Dataset} & \multirow{2}[2]{*}{$\texttt{ipc}$} & \multicolumn{3}{c}{ConvNet} & \multicolumn{2}{c}{ResNet-18} & \multicolumn{2}{c}{ResNet-50} & \multicolumn{2}{c}{ResNet-101}\\
	 \cmidrule(lr){3-5} \cmidrule(lr){6-7} \cmidrule(lr){8-9} \cmidrule(lr){10-11}
 & & MTT~\cite{MTT} & DataDAM~\cite{DataDAM} & TESLA~\cite{tesla} & SRe2L~\cite{sre} & DWA (ours) & SRe2L & DWA (ours) & SRe2L & DWA (ours)\\
	 \midrule
	 \multirow{2}{*}{Tiny-ImageNet} & $50$ & $\ApmB{28.0}{0.3}$ & $\ApmB{28.7}{0.3}$ & - & $\ApmB{41.1}{0.4}$ & \cellcolor{Color-p!90}$\ApmB{52.8}{0.2}$ & $\ApmB{42.2}{0.5}$ & \cellcolor{Color-p!90}$\ApmB{53.7}{0.2}$ & $\ApmB{42.5}{0.2}$ & \cellcolor{Color-p!90}$\ApmB{54.7}{0.3}$ \\
 & $100$ & - & - & - & $\ApmB{49.7}{0.3}$ & \cellcolor{Color-p!90}$\ApmB{56.0}{0.2}$ & $\ApmB{51.2}{0.4}$ & \cellcolor{Color-p!90}$\ApmB{56.9}{0.4}$ & $\ApmB{51.5}{0.3}$ & \cellcolor{Color-p!90}$\ApmB{57.4}{0.3}$ \\
	 \midrule
	 \multirow{3}{*}{ImageNet-1K} & 10 & $\ApmB{64.0}{1.3}^{\dag}$ & $\ApmB{6.3}{0.0}$
 & $\ApmB{17.8}{1.3}$ & $\ApmB{21.3}{0.6}$ & \cellcolor{Color-p!90}$\ApmB{37.9}{0.2}$ & $\ApmB{28.4}{0.1}$ & \cellcolor{Color-p!90}$\ApmB{43.0}{0.5}$ & $\ApmB{30.9}{0.1}$ & \cellcolor{Color-p!90}$\ApmB{46.9}{0.4}$\\
 & 50 & - & - & $\ApmB{27.9}{1.2}$ & $\ApmB{46.8}{0.2}$ & \cellcolor{Color-p!90}$\ApmB{55.2}{0.2}$ & $\ApmB{55.6}{0.3}$ & \cellcolor{Color-p!90}$\ApmB{62.3}{0.1}$ & $\ApmB{60.8}{0.5}$ & 
 \cellcolor{Color-p!90}$\ApmB{63.3}{0.7}$\\
 & 100 & - & - & - & $\ApmB{52.8}{0.3}$ & \cellcolor{Color-p!90}$\ApmB{59.2}{0.3}$ & $\ApmB{61.0}{0.4}$ & \cellcolor{Color-p!90}$\ApmB{65.7}{0.4}$ & $\ApmB{62.8}{0.2}$ & \cellcolor{Color-p!90}$\ApmB{66.7}{0.2}$\\
 % \midrule
	 \bottomrule 
	 \end{tabular}
% }
	 \label{tab:imagnet}
\end{table}

\begin{figure}[h] 
	 \centering
	 \includegraphics[width = 0.9\linewidth]{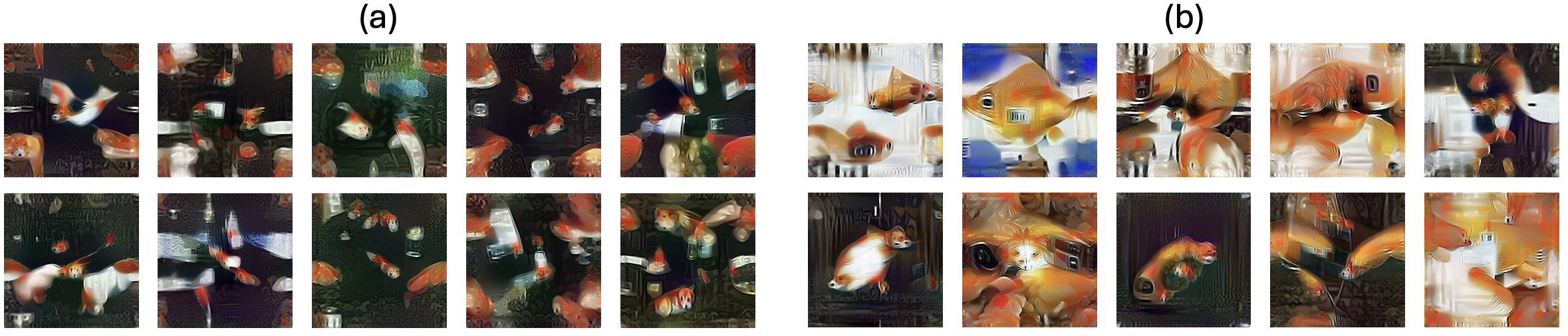}
	 \caption{\small Visualization of distilled images for the goldfish class. Panels (a) and (b) show the synthesized results by SRe2L~\cite{sre} and our DWA, respectively. The synthetic data instances generated by our DWA method exhibit significantly greater diversity compared to those produced by SRe2L, highlighting the effectiveness of our approach in capturing a broader range of features.}
	 \label{fig:visualization}
 %\vspace{-3pt}
\end{figure}
\begin{figure}[t]
	 \centering
	 \includegraphics[width = 0.9\linewidth]{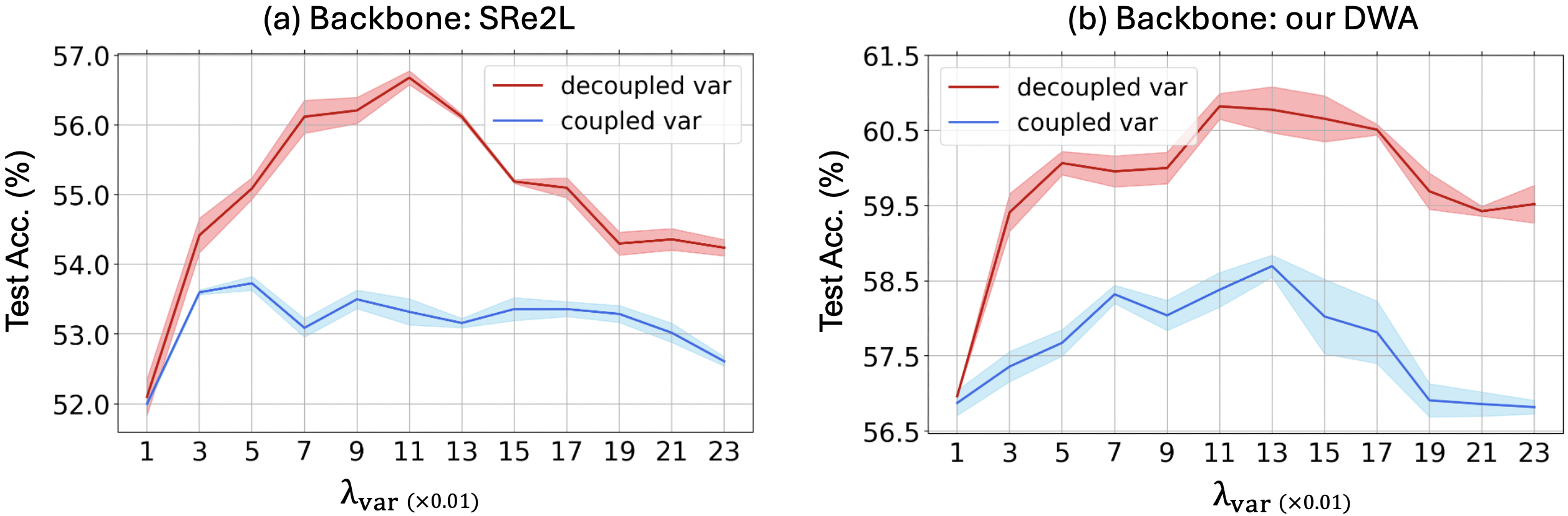}
	 \caption{\small Analysis of decoupled $\gL_\mathrm{var}$ coefficient. We vary $\lambda_\mathrm{var}$ across a wide range of $ (0.01 \sim 0.23)$. `decoupled var' indicates $\lambda_\mathrm{var}$ is changing individually with a fixed mean component whose weight defaults to 0.01. `coupled var' represents the weight of the mean and $\lambda_\mathrm{var}$ change in tandem. (a) and (b) illustrate the performance of the original SRe2L~\cite{sre} and our DWA in these two scenarios, respectively. This analysis is conducted on CIFAR-100 using ResNet-18. Each $\lambda_\mathrm{var}$ undergoes five independent experiments, with variance indicated by lighter color shades.}
	 \label{fig:Decoupled_Lvar}
  \vspace{-1em}
\end{figure}
\begin{wrapfigure}{r}{0.5\textwidth}
	 \centering
	 \vspace{-8pt}
	 \includegraphics[width = 0.45\textwidth]{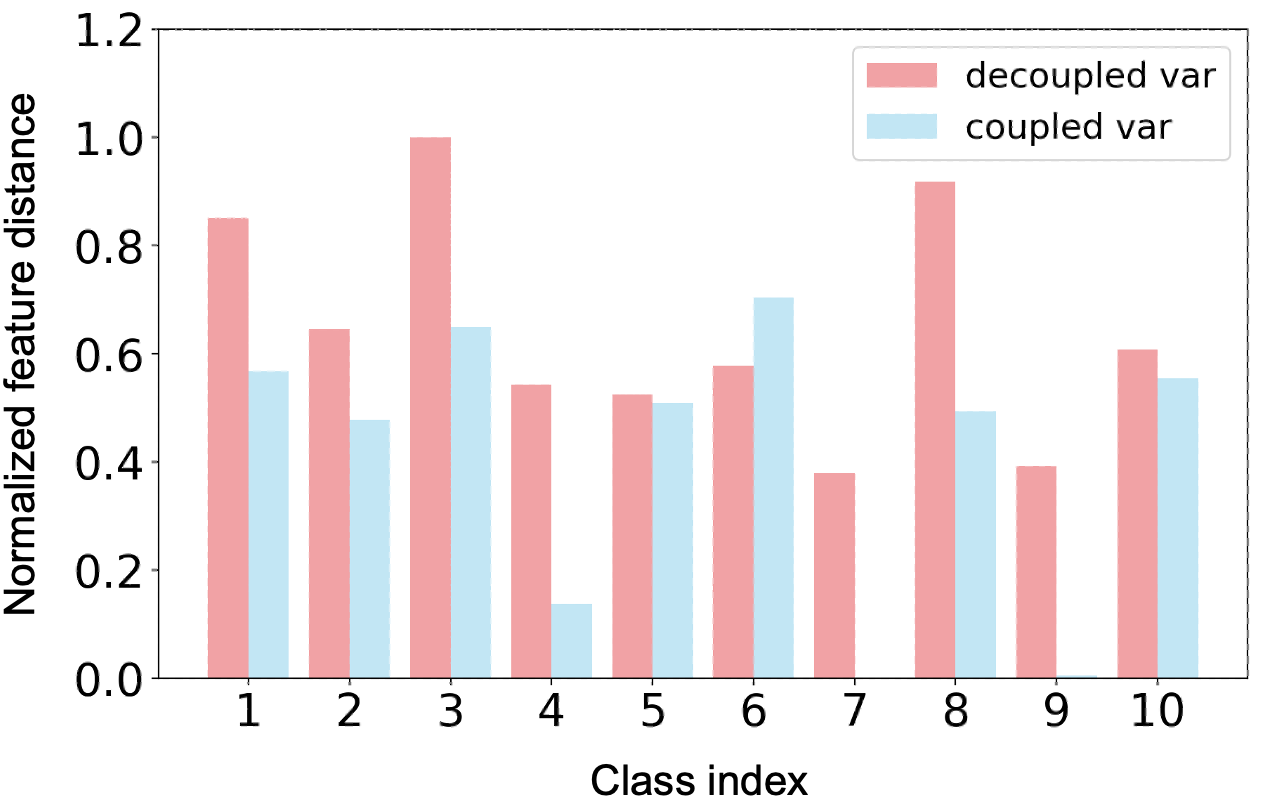} 
	 \caption{\small Normalized feature distance of decoupled variance component with $\lambda_\mathrm{var} = 0.11$ (the weight of mean component defaults to $0.01$) and coupled variance component with $\lambda_{\mathrm{BN}} = 0.11$. ResNet-18's last convolutional layer outputs are used for feature distance calculation (see \Cref{app:feature_dis}). Ten classes are randomly chosen from CIFAR-100 distilled dataset.} 
 % \vspace{-40pt}
	 \label{fig:feature_dis} 
\end{wrapfigure}

%\\ \noindent
\textbf{Tiny-ImageNet \& ImageNet-1K.}
Compared with CIFAR-10/100, ImageNet datasets are more closely reflective of real-world scenarios. \autoref{tab:imagnet} lists the related results. Due to the limited scalability capacity of conventional distillation paradigm, only a few methods have conducted evaluation on ImageNet datasets. Here we provide a comprehensive comparison with SRe2L~\cite{sre}, which has been validated as the most effective one for distilling large-scale dataset. It is obvious that our method significantly outperforms SRe2L on all $\texttt{ipc}$ settings and validation models. For instance, our DWA surpasses SRe2L by 16.6\% when $\texttt{ipc} = 10$ on ImageNet-1K using ResNet-18. \autoref{fig:visualization}
further provides the visualization results, the enhanced diversity is the key driver behind the substantial performance improvement.
% \vspace{-2.3em}
\subsection{Ablation Study}\label{sec:ablation}
\textbf{Decoupled $\gL_\mathrm{var}$ Coefficient.} 
%We first verify our hypothesis as stated in \Cref{sec: decoupled_varaince} that strengthening $gL_\mathrm{mean}$ contradicts the emphasis on $\gL_\mathrm{var}$, which is the primary factor to ensure the diversity in synthetic datasets. We use the normalized feature distance as the measurement of diversity, comparing the synthetic dataset distilled with the emphasis on $\lambda_{\mathrm{BN}}$ (strengthening both $gL_\mathrm{mean}$ and $gL_\mathrm{var}$) to the one with the emphasis on $gL_\mathrm{var}$, respectively. As illustrated in \autoref{fig:feature_dis}, the sole emphasis on $gL_\mathrm{var}$ achieves more diverse synthetic dataset in comparison the the same amount of emphasis on both $gL_\mathrm{mean}$ and $gL_\mathrm{var}$. This results confirm our hypothesis stated in \Cref{sec: decoupled_varaince}. 
We first test our hypothesis, as outlined in \Cref{sec: decoupled_varaince}, positing that strengthening $\gL_\mathrm{mean}$ conflicts with the emphasis on $\gL_\mathrm{var}$, which is critical for ensuring diversity in synthetic datasets. Therefore, we compare the synthetic dataset distilled with an emphasis on $\gL_{\mathrm{BN}}$ (which strengthens both $\gL_\mathrm{mean}$ and $\gL_\mathrm{var}$) against one that emphasizes $\gL_\mathrm{var}$ alone. As depicted in \autoref{fig:Decoupled_Lvar}, focusing solely on $\gL_\mathrm{var}$ outperforms the combined emphasis on $\lambda_{\mathrm{BN}}$ in both SRe2L~\cite{sre} and our proposed Directed Weight Adjustment (DWA). These experimental results verify our hypothesis in \Cref{sec: decoupled_varaince}, indicating the optimal value of the decoupled coefficient $\gL_\mathrm{var}$ is 0.11.
We also employ the normalized feature distance as a metric to comprehensively evaluate our emphasis. This metric measures the mutual feature distances between instances, as defined in \Cref{app:feature_dis}. By randomly selecting 10 classes from CIFAR-100, we calculate the normalized feature distances between synthetic datasets emphasized by the decoupled $\gL_\mathrm{var}$ and the coupled $\gL_{\mathrm{BN}}$. The findings, illustrated in \autoref{fig:feature_dis}, validate our hypothesis from a different perspective.
%In this section, we validate our conjecture in \Cref{sec: decoupled_varaince} suggesting that decoupling the variance component can improve the quality of the distilled dataset by enhancing diversity. As illustrated in \autoref{fig:Decoupled_Lvar}, for both the original SRe2L and our methods, the change of the coefficient of BN loss, $\lambda$, indeed contributes to some performance gain during the early growth phase (blue lines). However, due to the confrontation between the mean and variance components, the performance gain benefited from the unified adjustment cannot sustain with growing weight. In comparison, decoupling the $\gL_\mathrm{var}$ coefficient can provide significantly larger accuracy improvement. 
%Besides, as shown in \autoref{fig:Decoupled_Lvar}\textcolor{red}{ (b)}, our method further extend this phenomenon to a wider $\lambda_\mathrm{var}$ range, indicating a more robust parameter choice. We set $\lambda_\mathrm{var} = 0.11$ and maintain mean component as default for our DWA. \autoref{fig:feature_dis} clearly elucidates this phenomenon by illustrating the feature distance of distilled images within a single class. It's evident that by decoupling the variance coefficient, diversity is significantly enhanced. This improved diversity is crucial especially for distilled dataset containing only a few images per class.

\textbf{Directed Weight Adjustment.} 
We clarify the necessity of restricting the direction of weight adjustment in \Cref{sec:3.3}. To test its effectiveness, we apply a random $\Delta \theta$, sampled from a Gaussian Distribution, to $\theta_\gT$. As shown in \autoref{tab:direction_abla}, we assess synthetic datasets derived from three scenarios: no weight adjustment, random weight adjustment, and our directed weight adjustment (DWA) method, using the CIFAR-100 dataset. The results, examined across various architectures, underscore the importance of directing weight adjustments in distillation processes. Notably, we observe performance degradation in the synthetic dataset optimized with random weight adjustment at $\texttt{ipc} = 10$ compared to those without weight adjustment. This decline occurs because, at smaller $\texttt{ipc}$ values, the noise introduced by random weight adjustment outweighs the benefits of diversity. However, as the number of synthetic instances increases, diversity becomes more effective in capturing a broader range of features, leading to improved performance, as reflected at $\texttt{ipc} = 50$.
%when employing a larger distillation budget ($\texttt{ipc}$ = 50), models trained on synthetic datasets distilled using perturbed $\theta_\gT$ exhibit superior classification accuracy compared to their non-perturbed counterparts. However, this approach yields diminishing returns in the scenario with a smaller budget ($\texttt{ipc}$ = 10). For instance, random perturbation degrades the performance of ResNet-50 by $11.5\%$. This degradation stems from the 
%noise caused by undirected weight adjustment, thereby compromising the quality of the synthetic dataset. By directing the perturbations as per \autoref{eq:delta_theta}, substantial performance enhancements are observed in both scenarios and are generalizable to diverse network architectures.
\begin{table}
	 \centering
	 \caption{\small An ablation study of DWA was conducted using various network architectures. The synthetic dataset was distilled by ResNet-18 from the CIFAR-100 dataset. We use \ding{56} to denote the distilled dataset without weight adjustment, $\bigcirc$ to denote the distilled dataset with random weight adjustment, and \ding{52} to represent Directed Weight Adjustment (DWA).}
	 \setlength\tabcolsep{14pt} 
 	 \scriptsize
  \renewcommand{\arraystretch}{1.2}
 %{\fontsize{9}{9}\selectfont
 	 \begin{tabular}{lccc|ccc}
 	 \toprule 
 & \multicolumn{3}{c}{$\texttt{ipc} = 10$} & \multicolumn{3}{c}{$\texttt{ipc} = 50$}\\ 
 	 \cmidrule(lr){2-4} \cmidrule(lr){5-7}
 Perturbation & \ding{56} & $ \bigcirc$ & \ding{52} & \ding{56} & $ \bigcirc$ & \ding{52}\\ 
 	 \midrule
 ResNet-18 & $\ApmB{30.6}{0.7}$ & $\ApmB{14.9}{0.1}$ & {\cellcolor{Color-p!90}$\ApmB{39.6}{0.6}$} & $\ApmB{56.1}{0.4}$ & $\ApmB{56.2}{0.6}$ & {\cellcolor{Color-p!90}$\ApmB{60.3}{0.5}$}\\ 
 ResNet-50 & $\ApmB{26.5}{1.1}$ & $\ApmB{15.0}{0.2}$ & {\cellcolor{Color-p!90}$\ApmB{35.2}{0.7}$} & $\ApmB{55.7}{0.9}$ & $\ApmB{57.1}{0.5}$ & {\cellcolor{Color-p!90}$\ApmB{60.6}{0.8}$}\\ 
 MobileNetV2 & $\ApmB{18.2}{0.5}$ & $\ApmB{14.4}{1.2}$ & {\cellcolor{Color-p!90}$\ApmB{27.8}{0.7}$} & $\ApmB{46.9}{0.9}$ & $\ApmB{50.7}{0.6}$ & {\cellcolor{Color-p!90}$\ApmB{53.6}{0.2}$}\\ 
 ShuffleNet & $\ApmB{10.3}{0.7}$ & $\ApmB{10.7}{0.1}$ & {\cellcolor{Color-p!90}$\ApmB{19.4}{0.9}$} & $\ApmB{30.9}{1.1}$ & $\ApmB{39.1}{0.1}$ & {\cellcolor{Color-p!90}$\ApmB{41.7}{0.8}$}\\ 
 EfficientNet & $\ApmB{11.8}{0.4}$ & $\ApmB{11.1}{0.7}$ & {\cellcolor{Color-p!90}$\ApmB{20.2}{0.4}$} & $\ApmB{28.6}{1.0}$ & $\ApmB{38.8}{1.0}$ & {\cellcolor{Color-p!90}$\ApmB{40.7}{0.3}$}\\
 	 \bottomrule 
 	 \end{tabular}
 	 \label{tab:direction_abla}
 	 \vspace{-1em}
\end{table}
\begin{table}[h] 
\vspace{-0.5em}
	 \centering
	 \caption{\small Cross-architecture performance of distilled dataset of CIFAR-100 using ResNet-18 and ConvNet-128.} 
	 \setlength\tabcolsep{6pt} 
 % \renewcommand{\arraystretch}{1.3}
 % {\fontsize{9}{9}\selectfont
	 \scriptsize
	 \begin{tabular}{ccl|cccccc}
	 \toprule 
 % \midrule
 % \multirow{2}{*}{$\texttt{ipc}$} & \multirow{2}{*}{Methods} & \multicolumn{6}{c}{ResNet-18$\rightarrow$Different evaluation models}\\
 % & & MobileNetv2 & ShuffleNet & EfficientNet & VGG-16 & ResNet-50 & ConvNet-128\\
 & {$\texttt{ipc}$} & {Methods} & MobileNetv2 & ShuffleNet & EfficientNet & VGG-16 & ResNet-50 & ConvNet-128\\
 \midrule
 & & SRe2L & $\ApmB{16.1}{0.5}$ & $\ApmB{11.8}{0.7}$ & $\ApmB{11.1}{0.3}$ & $\ApmB{19.2}{0.2}$ & $\ApmB{22.4}{1.3}$ & $\ApmB{19.4}{0.2}$\\
 \rowcolor{Color-p!90}\cellcolor{white} & \multirow{-2}{*}{\cellcolor{white}10} & DWA (ours) & $\ApmB{27.8}{0.7}$ & $\ApmB{19.4}{0.9}$ & $\ApmB{20.2}{0.4}$ & $\ApmB{30.0}{0.5}$ & $\ApmB{35.2}{0.7}$ & $\ApmB{27.3}{0.3}$\\
 \cmidrule (l){2-9}
 & & SRe2L & $\ApmB{43.2}{0.2}$ & $\ApmB{27.5}{1.1}$ & $\ApmB{24.9}{1.7}$ & $\ApmB{40.4}{1.2}$ & $\ApmB{52.8}{0.7}$ & $\ApmB{19.4}{0.2}$\\
\rowcolor{Color-p!90}\multirow{-4}{*}{\cellcolor{white}ResNet-18} & \multirow{-2}{*}{\cellcolor{white}50} & DWA (ours) & $\ApmB{53.6}{0.2}$ & $\ApmB{41.7}{0.8}$ & $\ApmB{40.7}{0.3}$ & $\ApmB{51.6}{0.4}$ & $\ApmB{60.6}{0.8}$ & $\ApmB{37.0}{0.3}$\\
 \midrule 
 & & SRe2L & $\ApmB{28.7}{1.3}$ & $\ApmB{25.3}{0.4}$ & $\ApmB{18.0}{0.9}$ & $\ApmB{21.5}{1.6}$ & $\ApmB{41.8}{0.2}$ & -\\
 \rowcolor{Color-p!90}\cellcolor{white} & \multirow{-2}{*}{\cellcolor{white}10} & DWA (ours) & $\ApmB{37.3}{0.1}$ & $\ApmB{25.3}{0.4}$ & $\ApmB{24.5}{0.4}$ & $\ApmB{29.6}{1.3}$ & $\ApmB{47.1}{0.3}$ & $\ApmB{47.6}{0.4}$\\ 
 \cmidrule (l){2-9}
 & & SRe2L & $\ApmB{48.8}{0.4}$ & $\ApmB{49.3}{0.7}$ & $\ApmB{45.7}{0.8}$ & $\ApmB{38.9}{0.5}$ & $\ApmB{53.4}{0.5}$ & -\\
 \rowcolor{Color-p!90}\multirow{-4}{*}{\cellcolor{white}ConvNet-128} & \multirow{-2}{*}{\cellcolor{white}50} & DWA (ours) & $\ApmB{53.5}{0.3}$ & $\ApmB{44.37}{0.4}$ & $\ApmB{45.7}{0.8}$ & $\ApmB{38.9}{0.5}$ & $\ApmB{56.3}{0.3}$ & $\ApmB{59.0}{0.1}$\\ 
	 \bottomrule 
	 \end{tabular}
  	 %\vspace{-1em}
	 \label{tab:cross_arch}
\end{table}

\begin{wrapfigure}{r}{0.5\textwidth}
	 \centering
	 \includegraphics[width = 0.4\textwidth]{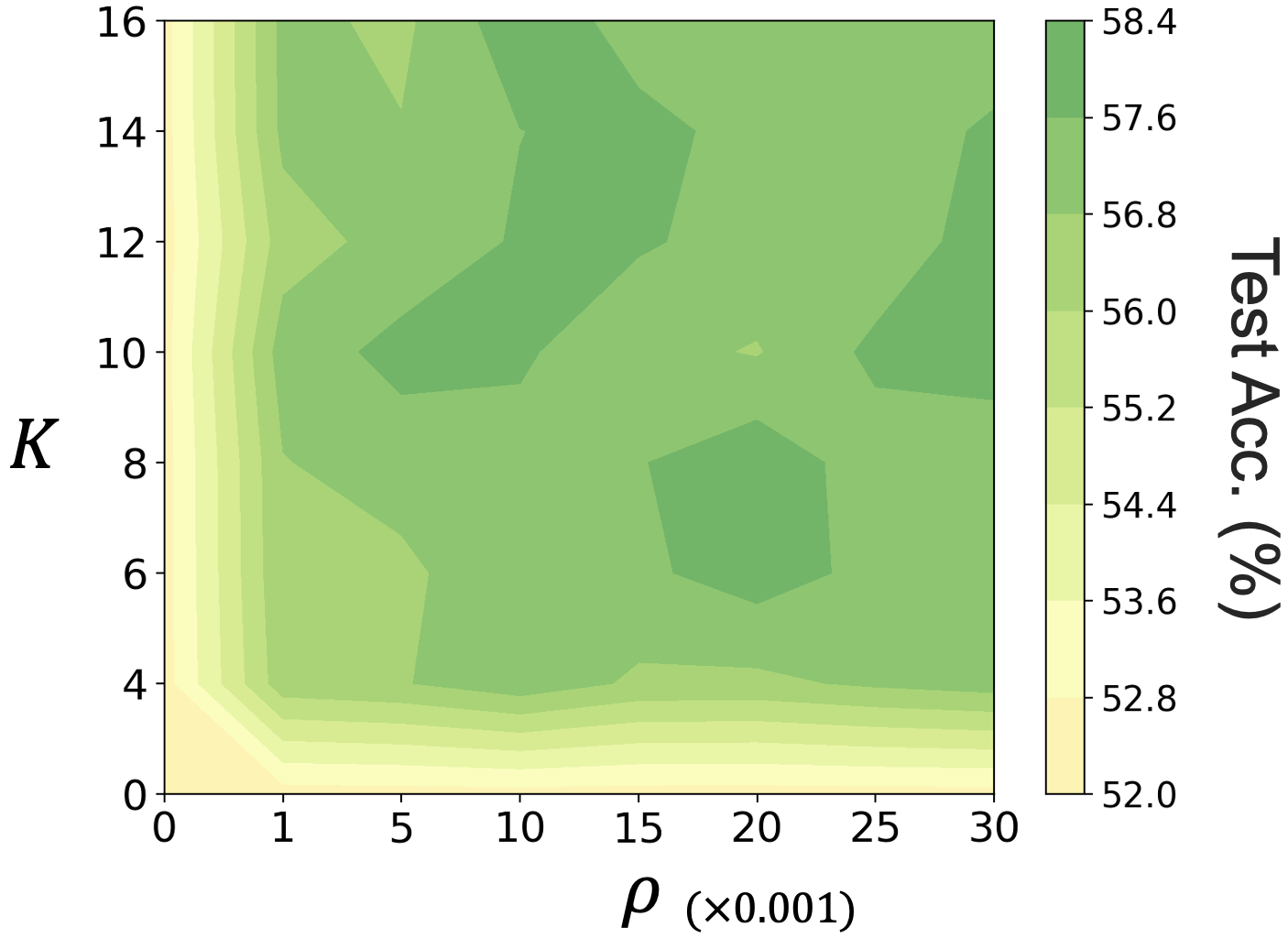} 
	 \vspace{-10pt}
	 \caption{\small Performance grid of ResNet-18 with changes in perturbation steps $K$ and magnitude $\rho$.} 
	 \label{fig:perturbation_hp} 
	 \vspace{-1em}
\end{wrapfigure}

\textbf{Parameters Study on $K$ and $\rho$.} Apart from direction, the number of steps $K$ and magnitude $\rho$ of perturbation also influence the distillation process. \autoref{fig:perturbation_hp} illustrates the grid search for these two hyper-parameters and demonstrates the positive impact of perturbation, which is achieved effortlessly, requiring no meticulous manual parameter tuning. In our experiments, we set $K = 12$ and $\rho = 15e^{-3}$ for all the datasets. Readers can adjust these hyper-parameters according to their specific circumstances (different datasets and networks) to obtain better results.

\textbf{Cross-Architecture Generalization.}
The generalizability across different architectures is a key feature for assessing the effectiveness of the distilled dataset. In this section, we evaluate the surrogate dataset condensed by different backbones (ResNet-18 and ConvNet-128) on various architectures including MobileNetV2~\cite{Mobilenetv2}, ShuffleNetV2~\cite{Shuffle}, EfficientNet-B0~\cite{efficientnet}, and VGGNet-16~\cite{vgg}. The experimental results are reported in \autoref{tab:cross_arch} and \autoref{tab:cross_arch_1}. It is evident that our DWA-synthesized dataset can effectively generalize across various architectures. Notably, for $\texttt{ipc} = 50$ on CIFAR-100 with ShuffleNetV2, EfficientNet-B0, and ConvNet-128—three architectures not involved in the data synthesis phase—our method achieves impressive classification performance, with accuracies of 41.7\%, 40.7\%, and 37.0\%, respectively, outperforming the latest SOTA method, SRe2L~\cite{sre}, by 14.2\%, 15.8\%, and 17.6\%. In \Cref{app:transformer_based_generalization}, we further extend the proposed method to a vision transformer-based model, DeiT-Tiny~\cite{deit}.
\begin{wraptable}{r}{0.5\textwidth}
 	 \centering
 \vspace{-10pt}
 	 \caption{\small Cross-architecture performance of distilled dataset of ImageNet-1K using ResNet-18.} 
 	 \vspace{5pt}
 	 \setlength\tabcolsep{5pt} 
 % \renewcommand{\arraystretch}{1.3}
 % {\fontsize{9}{9}\selectfont
 	 \scriptsize
 	 \begin{tabular}{cl|cccc}
 	 \toprule 
 % \midrule
 % \multirow{2}{*}{$\texttt{ipc}$} & \multirow{2}{*}{Methods} & \multicolumn{6}{c}{ResNet-18$\rightarrow$Different evaluation models}\\
 % & & MobileNetv2 & ShuffleNet & EfficientNet & VGG-16 & ResNet-50 & ConvNet-128\\
 {$\texttt{ipc}$} & {Methods} & MobileNetv2 & ShuffleNet & EfficientNet\\
 \midrule
 & SRe2L & $\ApmB{15.4}{0.2}$ & $\ApmB{9.0}{0.7}$ & $\ApmB{11.7}{0.2}$\\
 \rowcolor{Color-p!90}\multirow{-2}{*}{\cellcolor{white}10} & DWA (ours) & $\ApmB{29.1}{0.3}$ & $\ApmB{11.4}{0.6}$ & $\ApmB{37.4}{0.5}$\\ 
 \midrule
 & SRe2L & $\ApmB{48.3}{0.5}$ & $\ApmB{9.0}{0.6}$ & $\ApmB{53.6}{0.4}$\\
 \rowcolor{Color-p!90}\multirow{-2}{*}{\cellcolor{white}50} & DWA (ours) & $\ApmB{51.6}{0.5}$ & $\ApmB{28.5}{0.5}$ & $\ApmB{56.3}{0.4}$\\ 
 	 \bottomrule 
 	 \end{tabular}
 %\vspace{-6pt}
 	 \label{tab:cross_arch_1}
\end{wraptable}
\section{Related Works}
\label{sec:related}
Dataset Distillation~\cite{DD} emerges as a derivative of Knowledge Distillation (KD)~\cite{gou2021knowledge}, emphasizing data-centric efficiency over traditional model-centric one. Previous studies have explored various strategies to condense datasets, including performance matching, gradient matching~\cite{DC, zhao2021dataset, lee2022dataset}
distribution matching~\cite{CAFE, DM, zhao2023improved, zhang2023echo, deng2024exploiting}, and trajectory matching~\cite{MTT, tesla, FTD, du2024sequential,liu2024dataset,wang2024emphasizing}.

What distinguishes DD from KD is the bi-level optimization, which considers both model parameters and image pixels. The consequent complexity and computational burden intricate optimization significantly diminish the effectiveness of the aforementioned methods. To address this issue, {SRe2L}~\cite{sre} introduced a three-step paradigm known as \textit{Squeeze-Recover-Relabel}. This approach relies on the highly encoded distribution prior, \textit{i.e.}, the running mean and running variance in the BN layer, to circumvent supervision provided by model training. With this decoupled optimization, SRe2L is able to extend DD to high-resolution and large-scale datasets like ImageNet-1K.

Another critical challenge in dataset compression, not limited to distillation, is how to represent the original dataset distribution with a scarcity of synthetic data samples~\cite{sun2023diversity}. Previous research claims that the diversity of a dataset can be evaluated by spatial distribution~\cite{maharana2024mathbbd}, the maximum dispersion or convex hull volume~\cite{yu2022can}, and coverage~\cite{zheng2023coveragecentric}. Conventional dataset distillation~\cite{zhang2023accelerating, kim2022dataset} treats the synthetic compact dataset as an integrated optimizable tensor without specialized guarantees for diversity and relies entirely on the matching objectives mentioned above. Recognizing this limitation, Dream~\cite{liu2023dream} proposed using cluster centers to induce synthesis and ensure adequate diversity. 
Besides, SRe2L resorts to the second-order statistics, \textit{i.e.}, variance of representations in pre-trained weights to provide diversity. 
%However, due to the `one-by-one' distillation strategy, the absence of interation between instance leads to homogenized distillation results. 
%In this paper, we dedicate to enhance the diversity of distilled datatset.

\section{Conclusion}
In this work, we hypothesize that ensuring diversity is crucial for effective dataset distillation. Our findings indicate that the random initialization of synthetic data instances contributes minimally to ensuring that each instance captures unique knowledge from the original dataset. We validate our hypothesis through both theoretical and empirical approaches, demonstrating that enhancing diversity significantly benefits dataset distillation. To this end, we propose a novel method, Directed Weight Adjustment (DWA), which introduces diversity in synthesis by customizing weight adjustments for each mini-batch of synthetic data. This approach ensures that each mini-batch condenses a variety of knowledge. Extensive experiments, particularly on the large-scale ImageNet-1K dataset, confirm the superior performance of our proposed DWA method.

\textbf{Limitations and Future work.} While DWA provides a straightforward and efficient approach to introducing diversity in dataset distillation, its reliance on the sampling of a random distribution to adjust weight parameters presents limitations. Increasing the variance of the random distribution can introduce unexpected noise, thereby bottlenecking overall performance. Future investigations could explore synthesizing data instances in a sequential manner, encouraging later instances to consciously distinguish themselves from earlier ones, thereby further enhancing diversity.
\clearpage
\section*{Acknowledgements}
This research is supported by Jiawei Du's A*STAR Career Development Fund (CDF) C233312004 and Joey Tianyi Zhou’s A*STAR SERC Central Research Fund (Use-inspired Basic Research). This research is also supported by National Natural Science Foundation of China under Grant 62301213.

\bibliographystyle{plain} 
\bibliography{DWA.bib} % Specify the name of your .bib file (without the extension)

%%%%%%%%%%%%%%%%%%%%%%%%%%%%%%%%%%%%%%%%%%%%%%%%%%%%%%%%%%%%

\newpage
\appendix

\section{Appendix}
\subsection{Minimizing $\gL_\mathrm{mean}$ and $\gL_\mathrm{var}$ can be contradictory}\label{app:contradictory_mean_var}
To prove that minimizing $\gL_\mathrm{mean}$ and $\gL_\mathrm{var}$ can result in contradictory objectives for some existing instances, we will demonstrate that the gradients required to minimize $\gL_\mathrm{mean}$ and $\gL_\mathrm{var}$, respectively, may point in opposite directions. Specifically, for any arbitrary instance $\vs_i \in \gS$, our goal is to establish:
\begin{equation}
\label{eq:contra}
	\frac{\partial \gL_\mathrm{mean}}{\partial \vs_i} \cdot \frac{\partial \gL_\mathrm{var}}{\partial \vs_i} < 0, 
\end{equation}
For $\frac{\partial \gL_\mathrm{mean}}{\partial \vs_i}$, we have
\begin{align}
\label{eq:contra_mean}
	\frac{\partial \gL_\mathrm{mean}}{\partial \vs_i} & = \frac{\partial \left[\mu (\gS) - \mu \left(\gT \right) \right]^2}{\partial \vs_i} = \frac{\partial \left[\mu \left(\gS \right) - \mu \left(\gT \right) \right]^2}{\partial \mu \left(\gS \right)} \cdot \frac{\partial \mu \left(\gS \right)}{\partial \vs_i} \nonumber\\
	 & = 2 \left[\mu \left(\gS \right) - \mu \left(\gT \right) \right] \cdot \frac{1}{\left|\gS \right|}, 
\end{align}
because $\mu (\gS) = \frac{1}{|\gS|} \vs_i + \sum_{j \neq i} \frac{1}{|\gS|} \vs_j $, thus $\frac{\partial \mu (\gS)}{\partial \vs_i} = \frac{1}{|\gS|}$. For $\frac{\partial \gL_\mathrm{var}}{\partial \vs_i}$, we have 
\begin{align}
\label{eq:contra_var}
	\frac{\partial \gL_\mathrm{var}}{\partial \vs_i} & = \frac{\partial \left[\sigma^2 \left(\gS \right) - \sigma^2 \left(\gT \right) \right]^2}{\partial \vs_i} = \frac{\partial \left[\sigma^2 \left(\gS \right) -\sigma^2 \left(\gT \right) \right]^2}{\partial \sigma^2 \left(\gS \right)} \cdot \frac{\partial \sigma^2 \left(\gS \right)}{\partial \vs_i} \nonumber\\
	 & = 2 \left[\sigma^2 \left(\gS \right) - \sigma^2 \left(\gT \right) \right] \cdot \frac{\partial \sigma^2 \left(\gS \right)}{\partial \vs_i} \nonumber\\ 
	 & = 2 \left[\sigma^2 \left(\gS \right) - \sigma^2 \left(\gT \right) \right] \cdot \frac{\partial \left[\frac{1}{\left|\gS \right|} \left(\vs_i - \mu \left(\gS \right) \right)^2 + \sum_{j \neq i} \frac{1}{\left|\gS \right|} \left(\vs_j - \mu \left(\gS \right) \right)^2 \right] }{\partial \vs_i} \nonumber\\ 
	 & = 2 \left[\sigma^2 \left(\gS \right) - \sigma^2 \left(\gT \right) \right] \cdot \frac{1}{\left|\gS \right|} \frac{\partial \left(\vs_i - \mu \left(\gS \right) \right)^2}{\partial \vs_i} \nonumber\\ 
	 & = 2 \left[\sigma^2 \left(\gS \right) - \sigma^2 \left(\gT \right) \right] \cdot \frac{1}{\left|\gS \right|} \cdot 2 \left(\vs_i - \mu \left(\gS \right) \right) \cdot \frac{\partial \left(\vs_i - \mu \left(\gS \right) \right)}{\partial \vs_i} \nonumber\\ 
	 & = 2 \left[\sigma^2 \left(\gS \right) - \sigma^2 \left(\gT \right) \right] \cdot \frac{1}{\left|\gS \right|} \cdot 2 \left(\vs_i - \mu \left(\gS \right) \right) \cdot \left(1 - \frac{1}{\left|\gS \right|} \right).
\end{align}
Substitute \autoref{eq:contra_mean} and \autoref{eq:contra_var} back into \autoref{eq:contra}, 
\begin{align}
	 & \frac{\partial \gL_\mathrm{mean}}{\partial \vs_i} \cdot \frac{\partial \gL_\mathrm{var}}{\partial \vs_i} \nonumber\\
	 = & 2 \left[\mu \left(\gS \right) - \mu \left(\gT \right) \right] \cdot \frac{1}{\left|\gS \right|} \cdot 2 \left[\sigma^2 \left(\gS \right) - \sigma^2 \left(\gT \right) \right] \cdot \frac{1}{\left|\gS \right|} \cdot 2 (\vs_i - \mu (\gS)) \cdot (1 - \frac{1}{|\gS|}) \nonumber\\
	 = & \left[\frac {2}{\left|\gS \right|} \right]^3 \left(\left|\gS \right|-1 \right) \left[\mu \left(\gS \right) - \mu \left(\gT \right) \right] \cdot \left[\sigma^2 \left(\gS \right) - \sigma^2 \left(\gT \right) \right] \cdot \left(\vs_i - \mu \left(\gS \right) \right), 
\end{align}

% Obvisouly, the expected number of those instances would be around $\frac{1}{2}|\sS|$, because $\mu (\sS)$ is the mean of $\sS$.
Let $R = [\mu (\gS) - \mu (\gT)] \cdot [\sigma^2 (\gS) - \sigma^2 (\gT)] $, where $R $ is a constant that can be either positive or negative, depending on the values of $\mu (\gS), \mu (\gT), \sigma^2 (\gS) $, and $\sigma^2 (\gT) $. Suppose $R > 0 $. In this scenario, instances for which $(\vs_i - \mu (\gS)) < 0 $ will encounter contradictory objectives in optimization. Conversely, if $R < 0 $, instances where $(\vs_i - \mu (\gS)) > 0 $ will face similar contradictions.

\subsection{Experiments}
% \subsubsection{Comparison Baselines} 
% \label{app:baselines}
% \textbf{MTT} \cite{MTT} distills small datastet using trajectory matching-based strategy, and has achieved stasfictory performance on small dataset like CIFAR.\\
% \textbf{TESLA} \cite{tesla} scales MTT to large-scale dataset ImageNet-1K for the first time by restricting memory complexity to a constant.\\
% \textbf{DC} \cite{DM}:\\
% \textbf{CAFE} \cite{CAFE}:\\
% \textbf{FTD} \cite{FTD}:\\
\subsubsection{Hyper-parameter Settings} 
\label{app:para_setting}
\autoref{tab:hyper_c10}, \autoref{tab:hyper_tiny}, and \autoref{tab:hyper_img} list the hyper-parameter settings of our method on experimental datasets. We maintain consistency with SRe2L for a fair comparison.

\begin{table}[!t]
	 \centering
	 \caption{\small Hyper-parameter settings for CIFAR-10/100.}
	 \setlength\tabcolsep{4pt} 
	 \renewcommand{\arraystretch}{1.3}{\fontsize{9}{10}\selectfont
	 \begin{tabular}{cc|ccc}
	 \toprule
	 \multicolumn{2}{c|} {\textbf{Distillation}} & \multicolumn{2}{c}{\textbf{Validation}}\\ 
 	 \midrule
 \#Iteration & 1000 & \#Epoch & 400\\
 Batch Size & 100 & Batch Size & 128\\
 Optimizer & Adam with $\{\beta_1, \beta_2\} = \{0.5, 0.9\}$ & Optimizer & AdamW with weight decay of 0.01\\
 Learning Rate & 0.25 using cosine decay & Learning Rate & 0.001 using cosine decay &\\
 Augmentation & - & Augmentation & \renewcommand{\arraystretch}{0.9}{\begin{tabular}{c}
 \small \texttt{RandomCrop} \\
 \small \texttt{RandomHorizontalFlip}
 \end{tabular}}\\
 $\lambda_\mathrm{var}$ & 11 & Tempreture & 30\\
 $\rho, K$ & $15e^{-3}, 12$ &\\
 	 \bottomrule
	 \end{tabular}}
	 \label{tab:hyper_c10}
\end{table}

\begin{table}[!h] 
	 \centering
	 \caption{\small Hyper-parameter settings for Tiny-ImageNet.}\setlength\tabcolsep{6pt} \renewcommand{\arraystretch}{1.3}{\fontsize{9}{10}\selectfont
	 \begin{tabular}{cc|ccc}
	 \toprule
	 \multicolumn{2}{c|} {\textbf{Distillation}} & \multicolumn{2}{c}{\textbf{Validation}}\\ 
 	 \midrule
 \#Iteration & 2000 & \#Epoch & 200\\
 Batch Size & 100 & Batch Size & 128\\
 Optimizer & Adam with $\{\beta_1, \beta_2\} = \{0.5, 0.9\}$ & Optimizer & SGD with weight decay of 0.9\\
 Learning Rate & 0.1 using cosine decay & Learning Rate & 0.2 using cosine decay &\\
 Augmentation & \renewcommand{\arraystretch}{0.9}{\begin{tabular}{c}
 \small \texttt{RandomResizedCrop} \\
 \small \texttt{RandomHorizontalFlip}
 \end{tabular}} & Augmentation & \renewcommand{\arraystretch}{0.9}{\begin{tabular}{c}
 \small \texttt{RandomResizedCrop} \\
 \small \texttt{RandomHorizontalFlip}
 \end{tabular}}\\
 $\lambda_\mathrm{var}$ & 11 & Tempreture & 20\\
 $\rho, K$ & $15e^{-3}, 12$ &\\
 	 \bottomrule
	 \end{tabular}}
	 \label{tab:hyper_tiny}
\end{table}

\begin{table}[!h] 
	 \centering
	 \caption{\small Hyper-parameter settings for ImageNet-1K.}\setlength\tabcolsep{5.5pt} \renewcommand{\arraystretch}{1.6}{\fontsize{9}{10}\selectfont
	 \begin{tabular}{cc|ccc}
	 \toprule
	 \multicolumn{2}{c|} {\textbf{Distillation}} & \multicolumn{2}{c}{\textbf{Validation}}\\
 	 \midrule
 \#Iteration & 2000 & \#Epoch & 300\\
 Batch Size & 100 & Batch Size & 128\\
 Optimizer & Adam with $\{\beta_1, \beta_2\} = \{0.5, 0.9\}$ & Optimizer & AdamW with weight decay of 0.01\\
 Learning Rate & 0.25 using cosine decay & Learning Rate & 0.001 using cosine decay &\\
 Augmentation & \renewcommand{\arraystretch}{0.9}{\begin{tabular}{c}
 \small \texttt{RandomResizedCrop} \\
 \small \texttt{RandomHorizontalFlip}
 \end{tabular}} & Augmentation & \renewcommand{\arraystretch}{0.9}{\begin{tabular}{c}
 \small \texttt{RandomResizedCrop} \\
 \small \texttt{RandomHorizontalFlip}
 \end{tabular}}\\
 $\lambda_\mathrm{var}$ & 2 & Tempreture & 20\\
 $\rho, K$ & $15e^{-3}, 12$ &\\ 
	 \bottomrule
	 \end{tabular}}
	 \label{tab:hyper_img}
\end{table}
\subsubsection{Feature Distance Calculation} \label{app:feature_dis}
In \autoref{fig:feature_dis}, we use feature distance $\gD_{fea}$ to measure the diversity of distilled dataset. The following is how the class-wise feature distance is calculated, 
\begin{equation}
\gD_{fea}^{c} = \sum_{i = 1}^{\texttt{ipc}}\sum_{j = 1}^{\texttt{ipc}}\Vert g_{\theta_{\gT}} (\tilde{\vs}_i^c) - g_{\theta_{\gT}} (\tilde{\vs}_j^c) \Vert^{2}, 
\end{equation}
where $g_{\theta_{\gT}} (\tilde{\vs}_i^c)$ and $g_{\theta_{\gT}} (\tilde{\vs}_j^c)$ are the latent representations of $i$-th and $j$-th synthetic instances of class $c$, specifically the outputs from the last convolutional layer.
\subsubsection{Generalization to Vision Transformer-based Models} 
\label{app:transformer_based_generalization}
We acknowledge that our proposed approach cannot be directly applied to models without BN layers, such as Vision Transformers (ViTs). Our baseline solution, SRe2L, involves developing a ViT-BN model that replaces all LayerNorm layers with BN layers and adds additional BN layers between the two linear layers of the feed-forward network. We followed their solution and conducted cross-architecture experiments with DeiT-Tiny~\cite{deit} on the ImageNet-1K dataset. The results are listed in \autoref{tab:cross_arch_vit}. The results demonstrate that our approach can be applied to ViT-BN with superior performance compared to the baseline. 
% Theoretically, our approach can be combined with other generation-based distillation methods to avoid the BN limitation. 
\begin{table}[h] 
\vspace{-0.5em}
	 \centering
	 \caption{Generalization to a vision transformer-based model DeiT-Tiny.} 
	 % \setlength\tabcolsep{6pt} 
 % \renewcommand{\arraystretch}{1.3}
 % {\fontsize{10}{11}\selectfont
	 % \scriptsize
	 \begin{tabular}{cl|cccc}
	 \toprule 
 % \midrule
 % \multirow{2}{*}{$\texttt{ipc}$} & \multirow{2}{*}{Methods} & \multicolumn{6}{c}{ResNet-18$\rightarrow$Different evaluation models}\\
 % & & MobileNetv2 & ShuffleNet & EfficientNet & VGG-16 & ResNet-50 & ConvNet-128\\
  & {Methods} & DeiT-Tiny & ResNet-18 & ResNet-50 & ResNet-101\\
 \midrule
 &  SRe2L & $15.41$ & $46.80$ & $55.60$ & $60.81$\\
\rowcolor{Color-p!90}\multirow{-2}{*}{\cellcolor{white}ResNet-18} & DWA (ours) & $22.72$ & $55.20$ & $62.30$ & $63.3$\\
 \midrule
 &  SRe2L & $25.36$ & $24.69$ & $31.15$ & $33.16$ \\
\rowcolor{Color-p!90}\multirow{-2}{*}{\cellcolor{white}DeiT-Tiny-BN} &  DWA (ours) & $37.0$ & $32.64$ & $40.77$ & $43.15$ \\
	 \bottomrule 
	 \end{tabular}
  	 \vspace{-1em}
	 \label{tab:cross_arch_vit}
  % }
\end{table}
\subsubsection{Application to Downstream Tasks} 
We evaluate our proposed DWA on a continual learning task, based on an effective continual learning method GDumb~\cite{prabhu2020gdumb}. Class-incremental learning was performed under strict memory constraints on the CIFAR-100 dataset, with 20 images per class ($\texttt{ipc} = 20$). CIFAR-100 was divided into five tasks, and a ConvNet was trained on our distilled dataset, with accuracy measured as new classes were incrementally introduced. As shown in \autoref{tab:downstream}, DWA significantly outperforms SRe2L across all class-incremental stages, demonstrating superior retention of knowledge throughout the learning process.
\begin{table}[h]
    \caption{Application to continual learning task.}
    \centering
    \begin{tabular}{lccccc}\toprule
       Class  & $20$ & $40$ & $60$ &  $80$& $100$\\ \midrule
        SRe2L & $15.7$ & $10.6$ & $9.0$ & $7.9$ & $6.9$\\
        \rowcolor{Color-p!90} DWA (ours)& $34.6$ & $25.7$ & $22.5$ & $20.2$ & $18.1$\\ \bottomrule
    \end{tabular}
    \label{tab:downstream}
\end{table}
\subsubsection{Computational Overhead of Distillation}
We compare the average time required to generate one $\texttt{ipc}$ using ResNet-18 on CIFAR-100. As shown in \autoref{tab:overhead}, our proposed DWA incurs only a 7.32\% increase in computational overhead while significantly enhancing the diversity of the synthetic dataset.
This additional overhead arises from the $K$-step directed weight perturbation applied before generating each $\texttt{ipc}$, as detailed in lines 6-7 of \Cref{alg:dwa},
\begin{align*}
&\text{For } k = 1 \text{ to } K \text{ do} \quad & \\
&\quad \Delta\theta_k = \Delta\theta_{k-1} + \frac{\rho}{K} \nabla L_{\gS_{0}^i} \left( f_{\theta_T + \Delta\theta_{k-1}} \right).
\end{align*}
Since each $\texttt{ipc}$ requires $1000$ iterations of forward-backward propagation for generation, the additional $K = 12$ forward-backward propagations required by DWA are negligible in the overall distillation process.
\begin{table}[h]
    \centering
    \caption{Computational overhead of distillation on CIFAR-100 with ResNet-18.}
    \begin{tabular}{lc}
    \toprule
    {Methods} & {Avg. time for generating one $\texttt{ipc}$} \\ \midrule
    SRe2L & 116.58 s (100\%) \\
    \rowcolor{Color-p!90} DWA (ours) & 125.12 s (107.32\%) \\ \bottomrule
    \end{tabular}
    \label{tab:overhead}
\end{table}

% \begin{figure}[h]
% 	 \centering
% 	 \includegraphics[width = 1.0\linewidth]{figs/c100-visual.png}
% 	 \caption{\small Enter Caption}
% 	 \label{app:visualization}
% \end{figure}

\end{document}